\documentclass{article}

\usepackage[preprint]{neurips_2024}

\usepackage{epsfig}
\usepackage{enumitem}
\usepackage{bm}
\usepackage{makecell}
\usepackage{multirow}
\usepackage[ruled,vlined]{algorithm2e}
\usepackage{algorithmic}

\usepackage{mathtools} 
\usepackage{booktabs} 
\usepackage{tikz} 

\usepackage[utf8]{inputenc} 
\usepackage[T1]{fontenc}    
\usepackage{hyperref}       
\usepackage{url}            
\usepackage{booktabs}       
\usepackage{amsfonts}       
\usepackage{nicefrac}       
\usepackage{microtype}      
\usepackage{xcolor}         
\usepackage{makecell}
\usepackage[accsupp]{axessibility}  
\usepackage{makecell}
\usepackage{biblatex}
\usepackage{color}
\usepackage{pifont}

\usepackage{orcidlink}

\usepackage{graphicx}
\usepackage{authblk}

\title{Openstory++ : A Large-scale Dataset and Benchmark for Instance-aware Open-domain Visual Storytelling}

\author[1,2]{$^*$Zilyu Ye}
\author[1,2]{$^*$Jinxiu Liu}
\author[1,2]{$^*$Ruotian Peng}
\author[2]{JinJin Cao}
\author[2,4]{Zhiyang Chen}
\author[1]{Yiyang Zhang}
\author[3]{Ziwei Xuan}
\author[3]{Mingyuan Zhou}
\author[5]{Xiaoqian Shen}
\author[5]{Mohamed Elhoseiny}
\author[1]{$^\dagger$Qi Liu}
\author[2]{$^\dagger$Guo-Jun Qi}

\affil[1]{South China University of Technology}
\affil[2]{Westlake University}
\affil[3]{OPPO US Research Center}
\affil[4]{Foundation Model Research Center, CASIA}
\affil[5]{King Abdullah University of Science and Technology}

\begin{document}

\maketitle

\renewcommand{\thefootnote}{\fnsymbol{footnote}}

\footnotetext[1]{Equal contribution.}
\footnotetext[2]{Corresponding authors.}
\phantomsection
\addtocounter{footnote}{1}
\footnotetext{\{zilyuye,jinxiuliu0628\}@foxmail.com, prtprt666@gmail.com, drliuqi@scut.edu.cn, guojunq@gmail.com}

\renewcommand{\thefootnote}{\arabic{footnote}}

\begin{figure}[ht]
    \centering
    \includegraphics[width=1.02\textwidth]{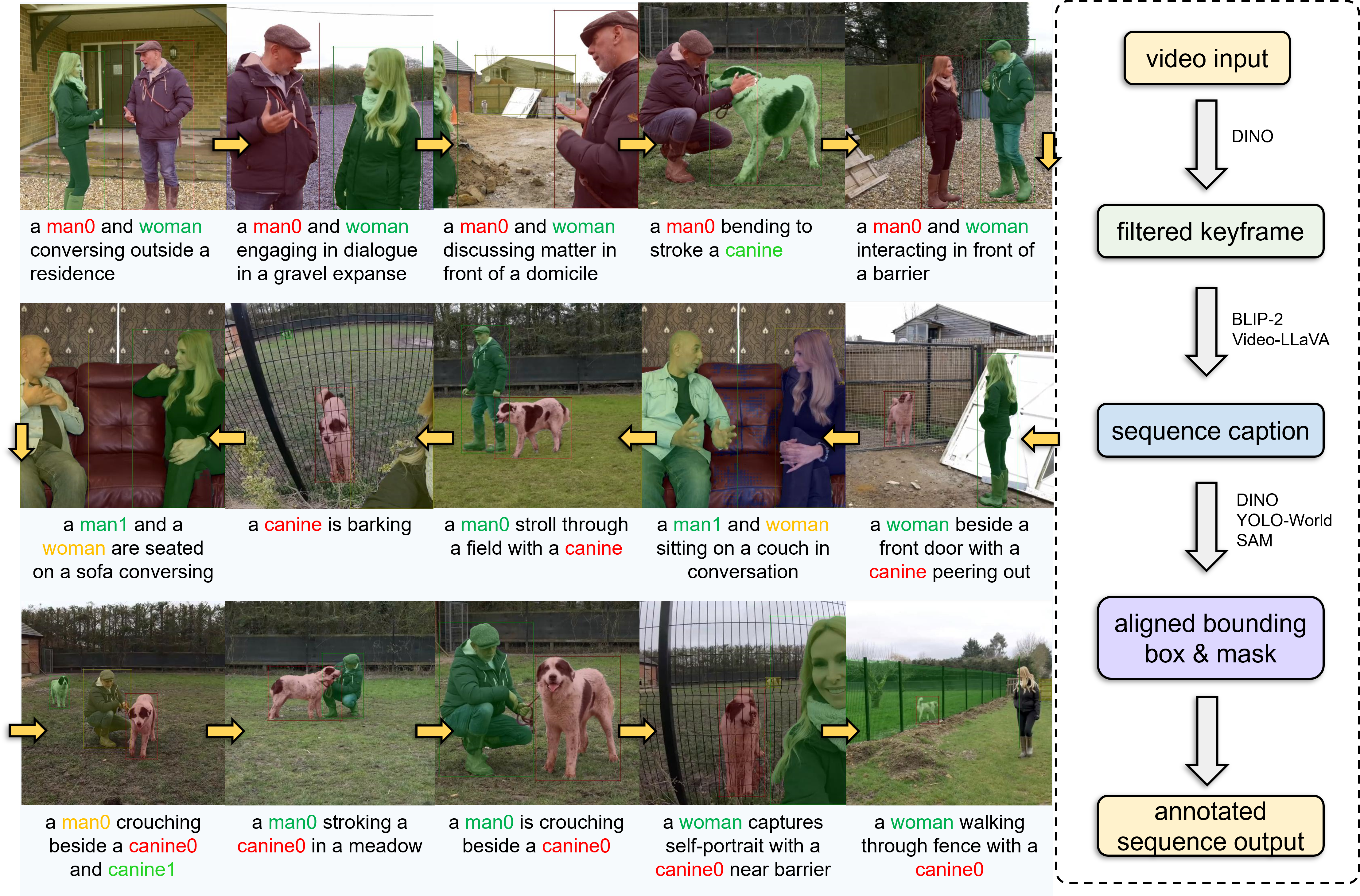}
    \caption{The visualization of our dataset. On the left is a data case with visual annotation that corresponds to each entity word in the sentence, where different color stands for different instance visual annotations, and on the right is the general pipeline of our dataset annotation process.}
    \label{fig:enter-label}
\end{figure}

\begin{abstract}

Recent image generation models excel at creating high-quality images from brief captions. However, they fail to maintain consistency of multiple instances across images when encountering lengthy contexts. This inconsistency is largely due to in existing training datasets the absence of granular instance feature labeling in existing training datasets. To tackle these issues, we introduce \textbf{Openstory++}, a large scale dataset combining additional instance-level annotations with both images and text. 
This dataset can be utilized to train multi-modal generated models, allowing for the training of instance-focused story visualization models.
Furthermore, we develop a tailored training methodology that emphasizes entity-centric image-text generation, ensuring that the models learn to effectively interweave visual and textual information. Specifically, Openstory++  streamlines the process of keyframe extraction from open-domain videos, employing vision-language models to generate captions that are then polished by a large language model for narrative continuity. It surpasses previous datasets by offering a more expansive open-domain resource, which incorporates automated captioning, high-resolution imagery tailored for instance count, and extensive frame sequences for temporal consistency. Additionally, we present \textbf{Cohere-Bench}, a pioneering benchmark framework for evaluating the image generation tasks when long multimodal context is provided, including the ability to keep the background, style, instances in the given context coherent. Compared to existing benchmarks,  our work fills critical gaps in multi-modal generation, propelling the development of models that can adeptly generate and interpret complex narratives in open-domain environments. Experiments conducted within Cohere-Bench confirm the superiority of Openstory++  in nurturing high-quality visual storytelling models, enhancing their ability to address sophisticated and open-domain generation tasks. More details can be found at \url{https://openstorypp.github.io/}

\end{abstract}

\section{Introduction}

The domain of artificial intelligence has witnessed a surge of interest due to the emergence of highly capable generative models. Modern Multi-modal Large Language Models (MLLMs) have achieved remarkable fluency in synthesizing text \cite{zeng2022glm, touvron2023llama, geminiteam2023gemini}, while state-of-the-art text-to-image models have shown an impressive ability to create realistic images \cite{stablediff, dalle3}. Despite these advancements, crafting high-quality visual stories that span an indefinite number of frames continues to be a significant challenge. Unlike the generation of a single image, creating multi-frame visual narratives necessitates the maintenance of subject continuity across frames, presenting a substantial hurdle.

While contemporary efforts have explored the image-text interleaved generation~\cite{liu2024visual, zhu2023minigpt, shen2023large, emu2, pan2023kosmos, ge2024seed, seed-llama, hu2024instruct, liu2024visual}, they have predominantly concentrated on sustaining general semantic and contextual relationships between text and images, falling short of achieving instance-level semantic coherence. Moreover, some studies~\cite{liu_intelligent_2024, li2019storyganandpororoSV, FlintstoneSV, lin2023videodirectorgptandCoref-SV} have only managed to maintain instance-level cross-frame visual consistency within confined domain datasets. This limitation is primarily due to the lack of open-domain datasets that encapsulate the narrative and temporal dynamics essential for training models to produce coherent stories. The existing datasets often fail to provide the required instance focus for coherent storytelling.

In response to these limitations and the pressing demand for automated and scalable methods to generate data for visual storytelling, we introduce \textbf{Openstory++}, a comprehensive dataset that underscores narrative continuity around pivotal instances with instance-level visual segmentation annotations, as depicted in Figure~\ref{fig:enter-label}. Our dataset processes video content to extract keyframes, evaluates them aesthetically, and employs BLIP2 \cite{li_blip-2_2023} to produce descriptive captions. These captions are further refined by a Large Language Model (LLM) to ensure narrative coherence. Additionally, our sub-pipeline identifies valid instances within the images and utilizes the Segment Anything Model (SAM) \cite{sam} to create masks for these instances. Openstory++ offers narratively coherent visual sequences that are tailored to specific instances, making it an ideal resource for training models that can generate visual stories with coherent instances across frames.

Additionally, to better reveal the model's ability in interleaved image-text generation, especially the consistency of visual appearance representing the same semantic instance across frames, we designed \textbf{Cohere-Bench}. It evaluates visual storytelling models across progressive dimensions, including single and multi-entity preservation in text-to-image generation and multi-turn generation capabilities, providing a comprehensive assessment of models' performance in visual storytelling tasks.

\begin{itemize}
    \item We created the Openstory++, a large-scale visual storytelling dataset enhancing instance-focused story visualization models by providing contextually coherent frames featuring recurring instances.
    \item We have developed a training methodology specifically tailored for entity-focused image-text interleaved generation, addressing the issue of inconsistency in existing models when dealing with lengthy contexts featuring multiple instances.
    \item We present Cohere-Bench, a pioneering benchmark framework for evaluating image-text generation. It overcomes the limitations of existing benchmarks by assessing long-context entity consistency and multi-turn generation capabilities.
    \item Experiments in \textbf{Cohere-Bench} demonstrate \textbf{Openstory++ }'s advantage in developing high-quality instance-aware open-domain visual storytelling models.
\end{itemize}

\section{Related Work}

\subsection{Datasets for Story Visualization}
Story visualization datasets are crucial for training models to generate images that correspond to narrative descriptions. However, existing datasets have several limitations. The VIST dataset\cite{VIST}, one of the earliest, did not include instance extraction within frames, which is essential for effective instance-focused story visualization tasks. More recent datasets like PororoSV\cite{li2019storyganandpororoSV}, FlintstonesSV\cite{FlintstoneSV}, and StorySalon\cite{liu_intelligent_2024} offer improved continuity in frame sequences and caption coherence but are limited to close-domain scenarios such as cartoons, restricting the model's capability for open-domain scene generation. Additionally, artificial annotations in captions of PororoSV, FlintstonesSV, and VIST limit the size of the datasets and increase production costs. On the other hand, datasets like DideMoSV\cite{maharana2022storydalle} that rely on video subtitles for caption generation may compromise the accuracy of captions in describing frame scenes. To overcome these limitations, we have developed a pipeline capable of extracting instance-focused keyframes and sequences from open-domain videos. This pipeline has been used to create the OpenStory++ dataset, a large-scale dataset that includes caption-frame pairs with long-term sequences. We believe that the OpenStory++ dataset will significantly enhance the proficiency of models in instance-focused story visualization tasks and provide a robust benchmark for multi-modal long-context understanding and generation.

\subsection{Benchmarks for Generative Multi-modal Model}
The advancement of multimodal large language models (MLLMs) has necessitated the development of benchmarks to evaluate their capabilities. Several studies have proposed benchmarks to assess various aspects of MLLMs. Benchmarks such as OwlEval\cite{ye2023mplug}, LLaVA-Bench\cite{liu2024visual}, LAMM\cite{gao2024lamm}, Touchstone\cite{bai2023touchstone}, MME\cite{fu2023mme}, and MMBench\cite{liu2023mmbench} evaluate the contextual understanding of text-image information and basic visual localization abilities of MLLMs, but they do not focus on generation capabilities. Other benchmarks, like MMMU\cite{yue2023mmmu}, emphasize the model's reasoning and cognitive abilities across multiple domains. Seed-bench-2\cite{seed-bench-2} specifically evaluates the image generation capabilities of multimodal large models. However, none of these benchmarks assess the thematic consistency of image generation in long-context scenarios for MLLMs. To address this gap, we introduce Cohere-Bench, a benchmark designed to evaluate the instance consistency of image generation in MLLMs over extended contexts.

\begin{table}[ht]
\centering
\setlength\tabcolsep{1pt}
\scalebox{0.99}{
\begin{tabular}{ccccccccc}
\toprule
Dataset & Domain & Caption & \thead{Frames} & \thead{Avg. \\ Length} \ & \thead{Masks \\ per frames} & Seriality  \\
\midrule
PororoSV\cite{li2019storyganandpororoSV}  &  close     & Manual & 73K & 5 &  - & \ding{51}            \\
FlintstonesSV\cite{FlintstoneSV}   &close& Manual & 123K & 5 &  - & \ding{51}               \\
DideMoSV\cite{maharana2022storydalle}&close& Manual & 53K & 3  & - & \ding{51}  \\
VIST\cite{VIST}     &open& Manual & 145K & 5 & -  & \ding{51}            \\
StorySalon\cite{liu_intelligent_2024} &close& ASR & 159K & 14  & 1 & \ding{51}             \\
SDD\cite{ma_subject-diffusionopen_2023} &open& Generated & 76M & 1  & 3 & \ding{55}             \\

Openstory++         & open & Generated & \textbf{100M+1M} & 28 & 2.5 & \ding{51}       \\
\bottomrule
\end{tabular}
}
\caption{Openstory++ statistics and comparison with other story-visualization datasets. Avg.Length represents the average length of storytelling data for each scene. Our dataset consists of about 100 million high-quality, fully annotated unique samples, as well as an additional 1 million fully annotated sequence samples.}
\label{tab:comparison}
\end{table}

\section{Openstory++ }
\label{sec:Openstory++ }
\begin{figure*}
    \centering
    \includegraphics[width=0.99\textwidth]{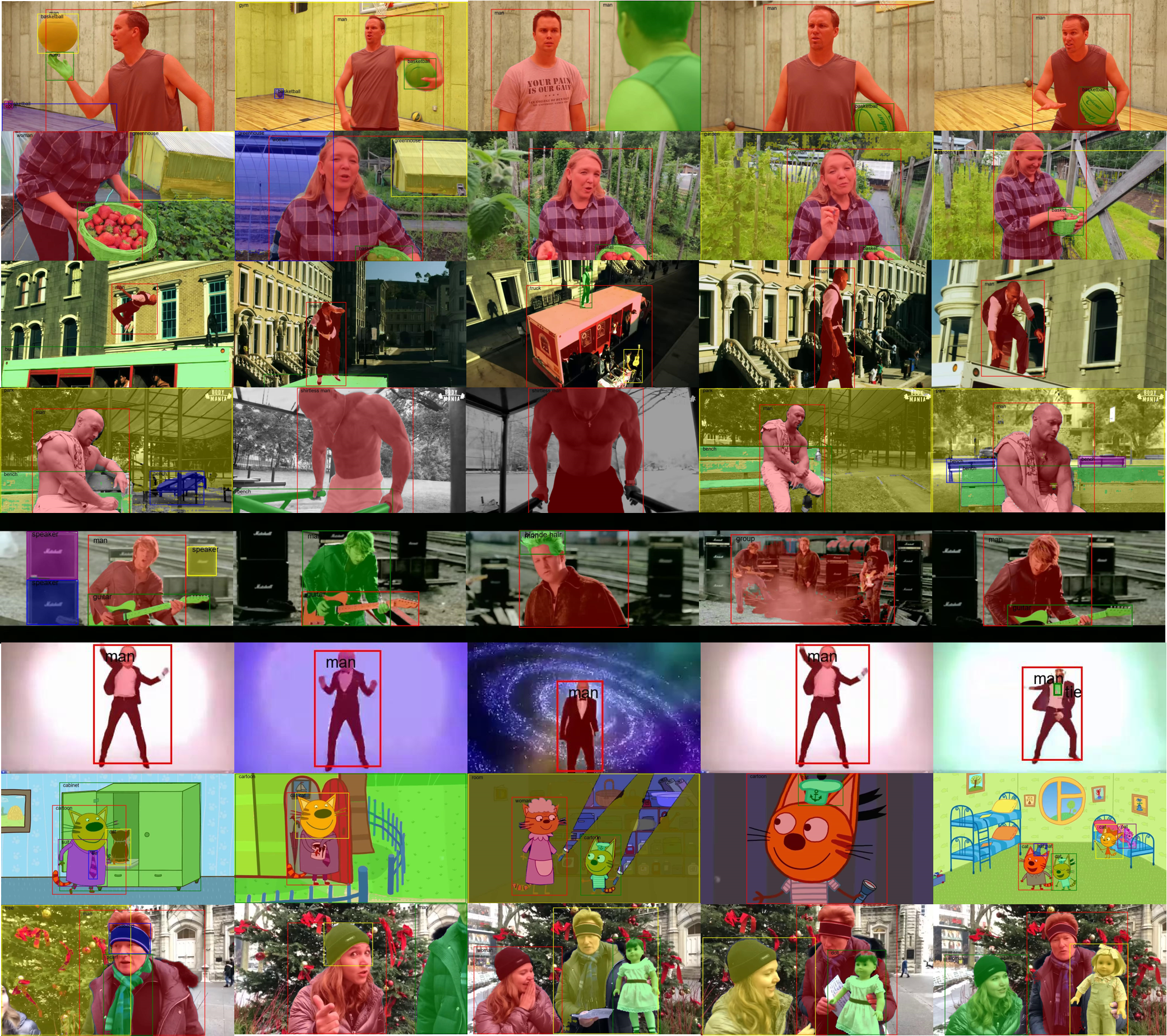}
    \caption{This figure shows the video frame sequence generated by our pipeline, with the subject's mask and the subject's bounding box.}
    \label{fig:resultshow}
\end{figure*}

Openstory++ is a dataset that enhances instance-focused story visualization models by providing contextually coherent frames featuring recurring instances, which promote the creation of coherent and contextually relevant visual narratives. Unlike previous approaches that generated a single keyframe per textual cue, our method extracts key frames emphasizing significant segments of the video, capturing the essential narrative aspects and offering a detailed analysis of instance activities. The result can be seen in Figure~\ref{fig:resultshow}.

\subsection{Data Sources}

Our sequence dataset incorporates a broad range of video datasets sourced from platforms such as Pandas-70M\cite{chen2024panda} and InternVid\cite{wang2023internvid}. Besides, we have carefully planned to add some high-quality data sources with inherent storytelling properties to enhance the data quality of the dataset. In contrast to the closed-domain datasets like StorySalon~\cite{liu_intelligent_2024} and Flintstones-SV~\cite{FlintstoneSV}, our dataset encompasses images from a wide array of scenarios, offering enhanced resources for training long-context models in open-domain environments. Using these data sources, we successfully labeled large-scale and fully annotated data. We compared other related datasets in Table~\ref{tab:comparison}.

\begin{figure*}
    \centering
    \includegraphics[width=1.05\textwidth]{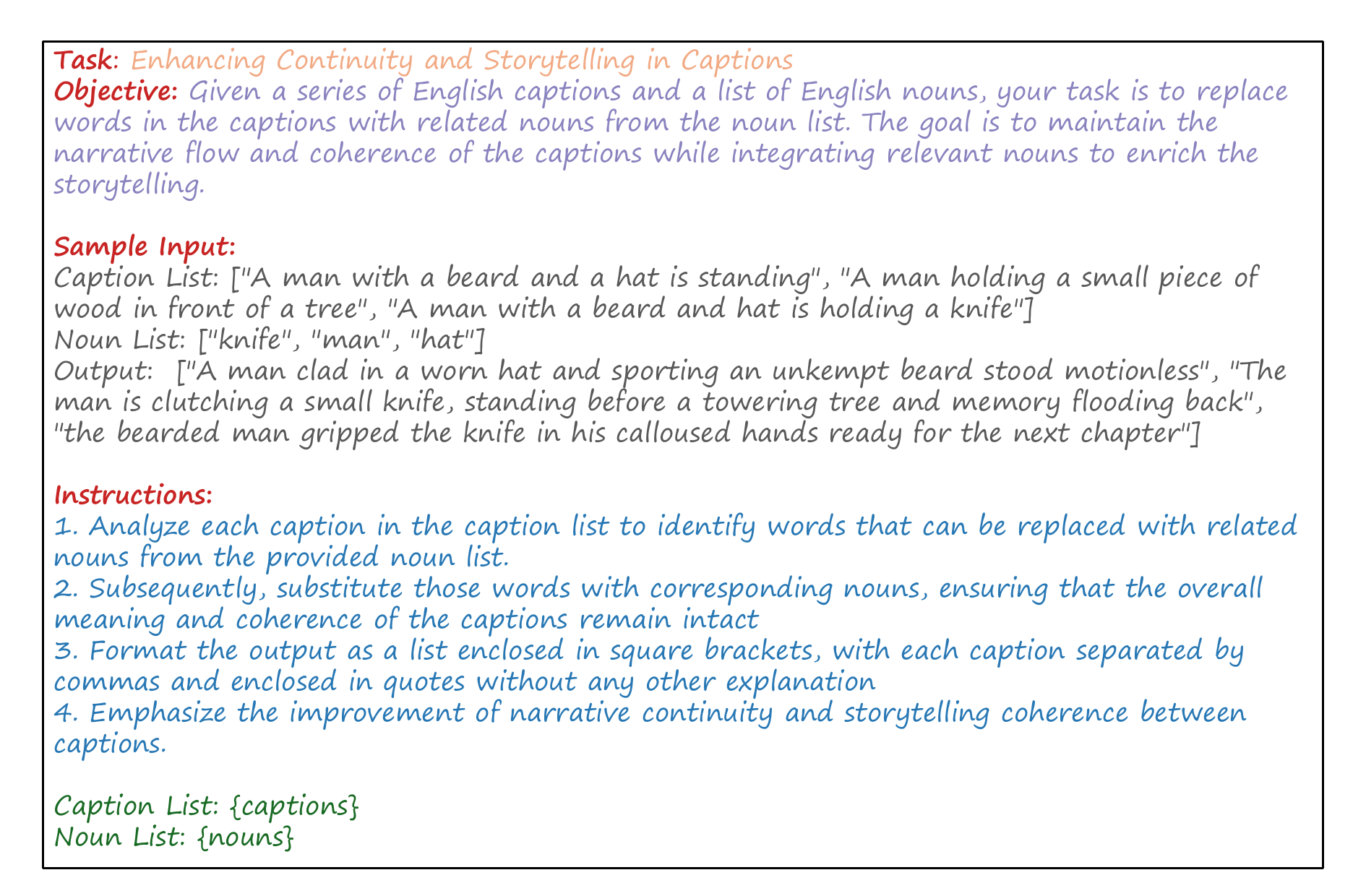}
    \caption{This figure presents a prompt designed to enhance narrative flow and coherence across scenes, which contains refined captioning guidelines aimed at enriching imagery with descriptive details while preserving the core content. Additionally, the prompt emphasizes maintaining consistent instances throughout the storytelling process.}
    \label{fig:prompt}
\end{figure*}

\subsection{Pipeline Overview}

\begin{figure*}
    \centering
    \includegraphics[width=0.99\textwidth]{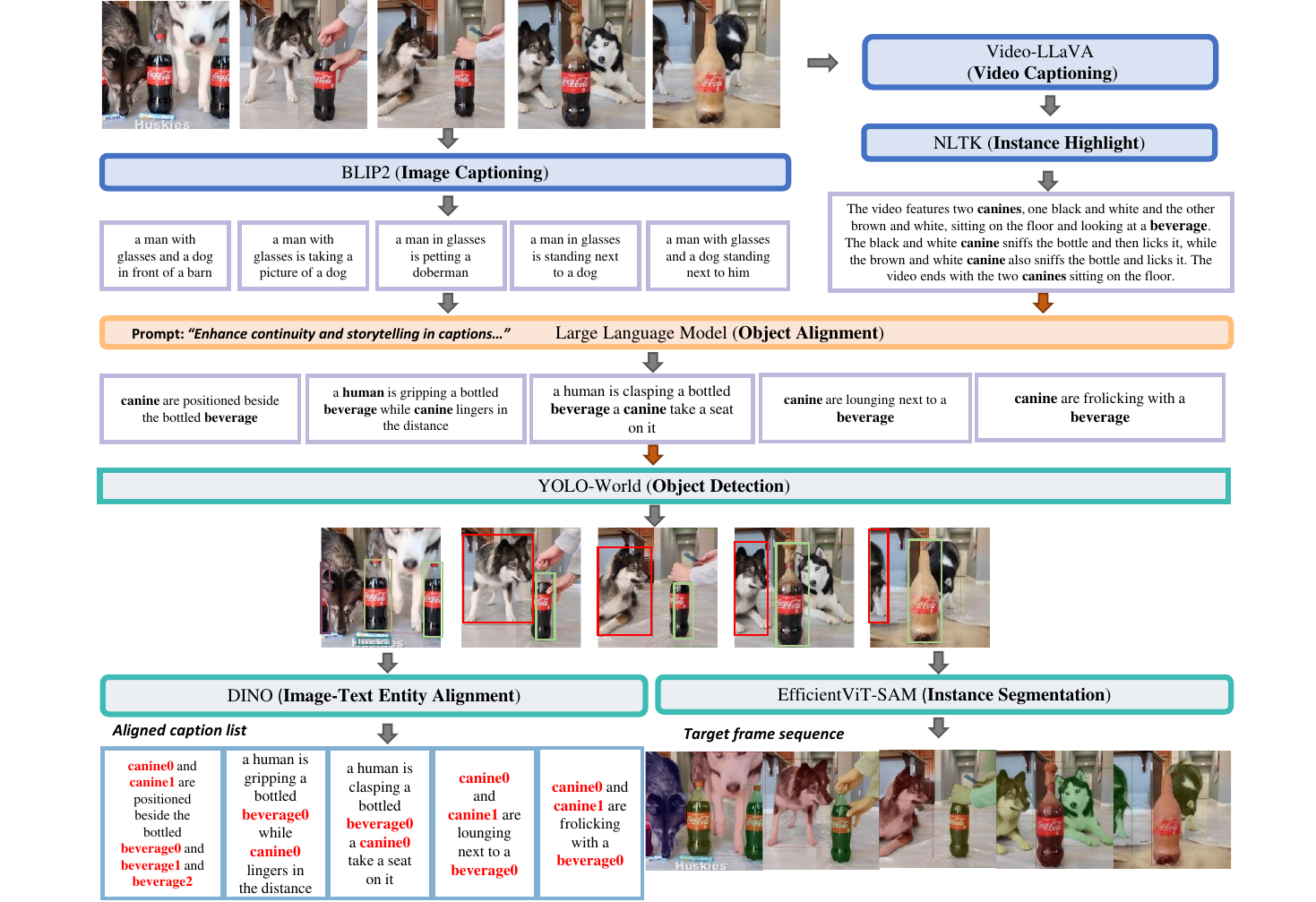}
    \caption{ This figure showcases the workflow of our pipeline. After obtaining a sequence of frames devoid of redundancy, we first utilized BLIP2 to generate basic image captions. Subsequently, Video-LLaVA was employed to produce a sequence of captions that encapsulate the narrative flow. Guided by the sequence caption, a  LLM was prompted to align the entities in the image captions, thus enhancing the narrative coherence across consecutive frames. Next, YOLO-World was applied to detect bounding boxes for the entities. To ensure that labels for the same entities across frames are unique and consistent, we blended the bounding box labels with the assistance of Dino and a facial feature module. Finally, we employed EfficientVIT-SAM to obtain the masks for the entities, thereby providing a comprehensive understanding of the spatial extent and characteristics of each entity within the frames. }
    \label{fig:generate_caption_mask}
    \vspace{0.5cm}
\end{figure*}

Our data processing pipeline is designed to construct a coherent series of keyframes featuring the same instance across different scenes, along with storytelling captions. This approach ensures narrative consistency and leverages a Large Language Model (LLM)~\cite{du2022glm} to enhance captions, maintaining coherence throughout the storytelling process. The pipeline culminates in creating precise instance masks from the refined keyframes and captions, essential for instance-focused visual storytelling tasks. The workflow is illustrated in Figure~\ref{fig:generate_caption_mask}.

\paragraph{Keyframe Extraction and Deduplication}
The pipeline initiates with extracting I-frames from the video content. Then, they are processed with DINOv2~\cite{oquab2023dinov2} to identify and eliminate redundant frames with high visual similarity. This ensures that each frame has distinct visual expression and captures the essence of the narrative.

\paragraph{Single-Image Captioning and Instance-Masking Workflow}
For each keyframe, we employ BLIP2~\cite{li2023blip} to autonomously generate basic captions. We subsequently utilize NLTK~\cite{bird2006nltk} to extract entity labels from these captions. Combined with YOLO-World~\cite{Cheng2024YOLOWorld}, we define the bounding box for each entity instance. The EfficientViT-SAM model~\cite{zhang2024efficientvit-SAM} then creates pixel-level instance masks, providing fine-grained visual annotations for each labeled entity in the captions. In addition, using this single-image annotation workflow, we can easily obtain fully annotated images from large-scale mature datasets\cite{laion400m, sun2024journeydb, sam}.

\paragraph{Frame-Caption Alignment for Narrative Coherence}
Then we align the refined captions with the keyframes to ensure consistent labeling of the same subject across images. VideoLLaVA~\cite{video-llava} processes multiple keyframes to generate sequence-level captions, enabling the standardization of entity labels across continuous scenes.  ChatGLM3-Turbo~\cite{zeng2022glm}, based on the Instance given by VideoLLaVA, use the prompt shown in Figure~\ref{fig:prompt} to refine the subjects represented by different concepts in captions annotated by BLIP2, ensuring that the subjects in multiple captions remain consistent, further refines individual image captions, focusing on narrative cohesion and enhancing the storytelling object alignment by accurately distinguishing between human instances.

\paragraph{Instance Masking}
The concluding step involves creating instance masks based on the labeled captions from the image sequence. With YOLO-World and DINOv2, we delineate the bounding box for each entity instance. The EfficientViT-SAM model fabricates pixel-level instance masks, concentrating our dataset on the most relevant visual storytelling instances. Additionally, an integrated facial detection mechanism enhances the precision of differentiating between human instances, overcoming the limitations of DINOv2.

\section{Training Data Challenge}
\subsection{Model Settings}
\label{Sec:model}

\begin{figure}
    \centering
    \includegraphics[width=0.95\textwidth]{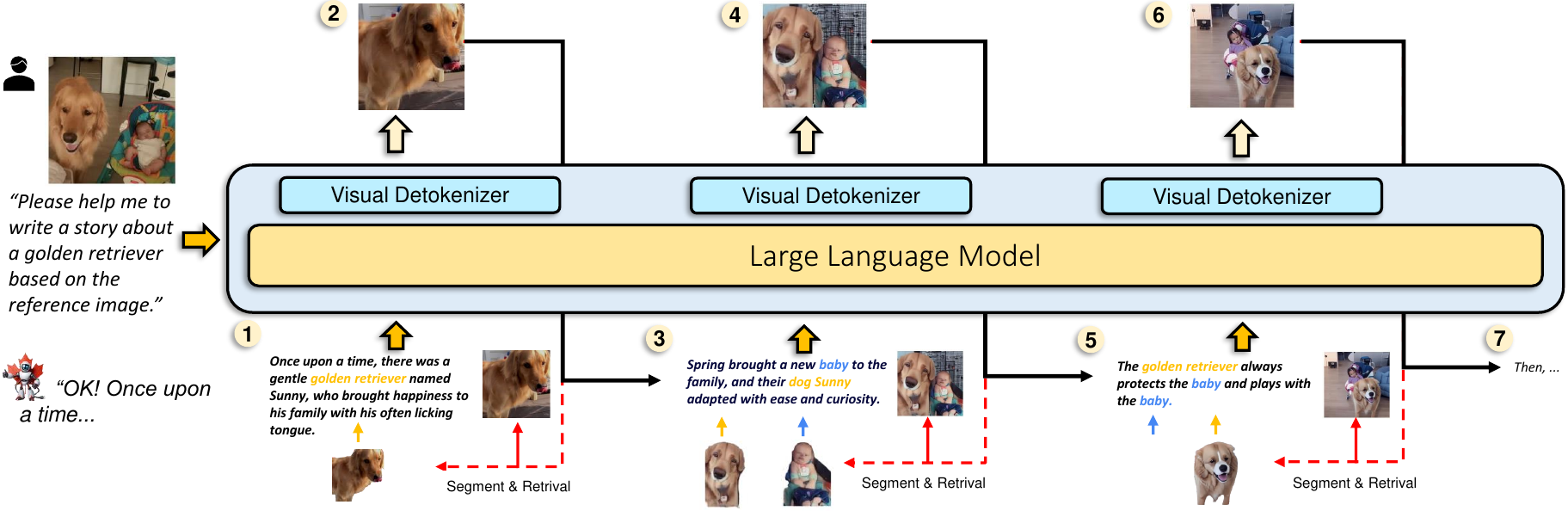}
    \caption{Overview of the interleaved image-text generation: Both the image and text are produced by the MLLM. During image generation, we take the diffusion model as the visual detokenizer.}
    \label{fig:model}
\end{figure}

In our experiment, we aim to generate $M$ images, denoted as $S_{\text{img}} = \{I^1, ..., I^M\}$, from $M$ lines of text prompts, $S_{\text{txt}} = \{L^1, ..., L^M\}$. Unlike traditional text-to-image models that generate images independently, our approach considers the contextual relationships and sequence of images. We use a multi-modal guided generation model that extends the autoregressive capabilities of MiniGPT-5. It is built on a Large Language Model (LLM) with pre-trained weights from Vicuna-7B~\cite{vicuna}, adept at producing consistent outputs from interleaved inputs of images and text. For visual encoding, we use a pretrained Vision Transformer (ViT)~\cite{dosovitskiy2020image} to convert visual content into embeddings and inject into LLM space by linear projection. The visual encoder extracts features from both original images and segmented objects, with segmented images encoded by a visual tokenizer. During image generation, the visual encoder and LLM process input images and text together. A visual de-tokenizer, which is Stable Diffuiosn 2.1~\cite{rombach2022high}, uses the LLM's output features to condition the final image generation. The architecture of model is shown in Figure~\ref{fig:model}.

\subsection{Dataset Comparsion}

\begin{table}[ht]
\centering
\setlength\tabcolsep{1pt}
\scalebox{0.99}{
\begin{tabular}{ccccccccccc}
\toprule
\thead{Training \\ Data} & \thead{Semantic \\ Alignment}$\uparrow$ & \thead{Style \\ Consistency}$\uparrow$ & \thead{Instance \\ Consistency}$\uparrow$ &  Perplexity$\downarrow$ & Aesthetic$\uparrow$  \\
\midrule
VIST~\cite{VIST}  &  0.183       &    0.742    &     0.598    & 35.626&4.748             \\
StorySalon~\cite{liu_intelligent_2024}   &0.245&0.754&0.727& 31.256& \textbf{4.941}              \\
FlintstoneSV~\cite{FlintstoneSV} &0.252&0.772&0.720& 30.254&4.864             \\
PororoSV~\cite{li2019storyganandpororoSV}     &0.247&0.662&0.637& 52.703& 4.755             \\
Openstory++         & \textbf{0.262} & \textbf{0.783} & \textbf{0.765} & \textbf{29.045} & 4.843         \\
\bottomrule
\end{tabular}
}
\caption{Comparison of various metrics across different datasets.}
\label{tab:metrics_comparison}
\end{table}

We evaluated our model across various datasets for its ability to generate coherent and stylistically consistent text, as detailed in Section~\ref{Sec:model}. Key metrics included semantic alignment, style consistency, and instance consistency. Perplexity, using phi3-mini-4k~\cite{abdin2024phi}, assessed word prediction accuracy. Aesthetic judgment evaluated visual appeal and distribution similarity to real images, respectively. From the Table~\ref{tab:metrics_comparison}, it is evident that our model trained on Openstory++ outperforms other datasets in terms of semantic alignment, style consistency, and instance consistency, as indicated by the higher values in these metrics. The lower perplexity score for our dataset suggests that our model is more effective at predicting the next word in a sequence. The aesthetic scores further validate the quality of our model's generated content, with a higher aesthetic score indicating better visual appeal except compared to close domain cartoon datasets. This demonstrates the crucial role of our dataset in generating high quality image-text interleaved models with cross-frame instance consistency.

\subsection{Cohere-Bench}
\label{sec:cohere-bench}
Our Cohere-Bench is designed to evaluate the quality of generated images and text against a ground truth dataset. We have sampled a subset of our dataset with 1600 items to serve as this evaluation dataset. The benchmark assesses the similarity of generated images to the ground truth and the effectiveness of the generated text. To evaluate text effectiveness, we employ a pipeline that involves extracting keyframes from videos, tagging them with Large Language Models (LLMs) and Video LLaVA~\cite{video-llava}, and applying YOLO-World~\cite{Cheng2024YOLOWorld} and BLIP2~\cite{li2023blip} to the generated images to add captions and identify the main instance's position, which is then aligned using LLM. This process results in a sequence of generated images with captions and instance-level segmentation. For consistency of the generated instance with the given instance, we measure semantic similarity between the segmented instance and the ground truth. We also calculate the similarity between the instance and previously appearing instances, as well as semantic alignment with the current instruction, style consistency, and instance consistency for both single-instance (s) and multiple-instance (m) settings. The specific calculation methods are as follows:

\begin{itemize}
    \item \textbf{Semantic Alignment:} We use CLIP~\cite{radford2021learning} to measure the semantic similarity between visual and text features, ensuring that the generated image aligns with the instruction prompt.
    \item \textbf{Background Consistency:} We detect entities using YOLO-World~\cite{Cheng2024YOLOWorld} and calculate the similarity of the masked and inpainted images to ensure background consistency.
    \item \textbf{Style Consistency:} We measure the similarity across previous frames and compute the average similarity score to gauge style consistency.
    \item \textbf{Instance Consistency:} For both single and multiple instances, we calculate the similarity between the generated image and the reference instance using YOLO-World~\cite{Cheng2024YOLOWorld} to segment the instance.
    \item \textbf{Instance Integrity:} We assess the completeness of the segmented instance in the current image compared to the ground truth, expressed as a percentage.
    \item \textbf{BLEU4:} 
    Due to the challenges in evaluating the accuracy of text continuation, we apply our dataset's annotation pipeline directly to the generated image sequence. By doing so, we annotate the generated images and use BLEU4~\cite{papineni2002bleu} scores to measure the similarity between the generated captions and the ground truth text. This approach allows us to assess the accuracy of content generation based on context, which measures the similarity of the recaptioned text to the ground truth. 
\end{itemize}

\paragraph{Evaluation analysis} 
 We evaluate the semantic alignment of generated images with their current captions and their visual consistency within a multi-modal context. The results are presented in Table~\ref{comparsion_quantivate}. For text alignment, our model, our model, outperforms all others except for GPT-4V and MiniGemini, which also demonstrate strong performance due to their robust captioning capabilities. Additionally, our model excels in maintaining cross-plot background consistency and instance consistency in both single and multiple instance settings. Unlike other models that rely on global features to encode images into the MLLM, resulting in less detailed visual features, our model enriches the inference phase with detailed, instance-level visual annotations. This approach integrates instance-level segmentation into the autoregressive space of the LLM, enhancing the model's ability to understand and generate detailed visual features.

\begin{table}[ht]
    \centering
    \setlength{\tabcolsep}{1pt} 
    \scalebox{0.90}{ 
    \begin{tabular}{@{}cccccccc@{}} 
        \toprule
        \textbf{Models} & 
        \thead{Semantic \\ Alignment}$\uparrow$ & 
        \thead{Background \\ Consistency}$\uparrow$ & 
        \thead{Style \\ Consistency}$\uparrow$ & 
        \thead{Instance \\ Consistency (s)}$\uparrow$ & 
        \thead{Instance \\ Consistency (m)}$\uparrow$ & 
        \thead{Instance \\ Integrity}$\uparrow$ & 
        BLEU4$\uparrow$ \\
        \midrule
        DreamLLM\cite{dong2023dreamllm} & {0.270}  & {0.615}  & {0.615}  & {0.271}  & 0.292 & 0.144 & 0.055 \\
        MiniGPT-5\cite{minigpt5} & {0.209}  & {0.634}  & {0.214}  & 0.214 & 0.219 & 0.115 & {0.011}  \\
        SEED-X\cite{seed-x} & {0.272}  & {0.775}  & {0.762}  & 0.744 & 0.774 & 0.421 & {0.057}  \\
        Emu2\cite{emu2} & {0.258}  & {0.788}  & {0.762}  & 0.818 & \textbf{0.787} & 0.351 & {0.058}  \\
        GPT4-V\cite{gpt4} & \textbf{0.286} & 0.762 & 0.781 & 0.753 & 0.761 & 0.424 & 0.062 \\
        MiniGemini\cite{minigemini} & {0.271}  & {0.710}  & {0.577}  & 0.602 & 0.610 & 0.203 & {0.052}  \\
        \thead{Our model \\ (w/o visual anno)} & 0.254 & 0.748 & 0.766 & 0.693 & 0.696 & 0.383 & 0.054 \\
        \thead{Our model \\ (w/ visual anno)} & 0.279 & \textbf{0.791} & \textbf{0.784} & \textbf{0.821} & 0.782 & \textbf{0.429} & \textbf{0.064} \\
        \bottomrule
    \end{tabular}
    }
    \caption{Quantitative comparison of various models on semantic alignment, style consistency, and instance consistency in single-instance (s) and multiple-instance (m) settings. The table also presents BLEU4 scores, which are metrics for evaluating the quality of generated text.}
    \label{comparsion_quantivate}
\end{table}

\subsection{Human Evaluation}
Our human evaluation framework integrates assessments of visual story quality across various dimensions, including image-text alignment, image style, story consistency, and instance consistency. We employ two complementary evaluation methods:

 \paragraph{Comparative Evaluation:} This method quantitatively measures the visual story quality through numerical ratings. Participants rate random samples from our model and baseline models on a scale of 1 to 5, with 1 being the lowest and 5 the highest quality. This approach provides detailed insights into each model's performance across different aspects of visual storytelling.
 \paragraph{Preference Evaluation:} This method captures the instanceive preferences of participants regarding the overall appeal of visual stories. For a given storyline, participants choose their most preferred image sequence from those generated by our model and other methods. This reveals which model's storytelling is most engaging to the audience.

The evaluation criteria, including semantic alignment, image style, story consistency, character consistency, plot continuity, and image quality, are chosen to reflect the key elements of visual storytelling. As shown in Table~\ref{tab:human_eval}, our model trained on Openstory++ (w/ visual anno), shows a strong performance across most metrics, indicating its effectiveness in generating coherent and high-quality visual narratives. However, it slightly underperforms in the text alignment metric compared to GPT4-V and MiniGemini. This discrepancy could be attributed to the advanced language understanding and visual generation capabilities of GPT4-V and the sophisticated image synthesis techniques provided by MiniGemini's training with DALL-E 3's API~\cite{dalle3}. Despite this, our model trained on Openstory++ (w/ visual anno) excels in plot continuity and image quality, which are crucial for maintaining the narrative flow and visual appeal of the stories. This suggests that training on datasets with visual annotations significantly enhances the model's ability to generate images that are not only consistent with the story context but also contribute to a coherent and engaging visual storytelling experience. 

\begin{table}[ht]
\centering
\setlength{\tabcolsep}{3pt}
\scalebox{0.99}{
\begin{tabular}{ccccccc|c}
\hline
\textbf{Model / Training Data} & \textbf{Align} & \textbf{Style} & \textbf{Consist} & \textbf{Char} & \textbf{\thead{Plot \\ Continuity}} & \textbf{\thead{Image \\ Quality}} & \textbf{Pref} \\
\hline
SEED-LLAMA~\cite{seed-llama} & 2.87 & 2.58 & 2.63 & 2.95 & 3.55 & 3.57 & 3.47\% \\
SEED-X~\cite{seed-x} & 3.76 & 3.78& 3.67 & 3.67 & 3.66& 3.93 & 13.53\% \\
Emu2~\cite{emu2} & 3.20 & 2.92 & 3.85 & 2.75 & 3.70 & 4.11 & 14.21\% \\
MiniGemini~\cite{minigemini} & 4.03& 3.81 & 2.85 & 2.70 & 2.65 & 3.62& 8.34\%\\
GPT4-V~\cite{gpt4} & \textbf{4.30} & 2.87 & 3.85 & 3.73& 4.25 & \textbf{4.75} & 20.66\% \\
VIST~\cite{VIST} & 1.98& 2.29 & 3.25 & 2.15 & 2.31 & 2.72 & 1.56\% \\
StorySalon~\cite{liu_intelligent_2024} & 3.95& 3.30 & 4.07& 3.22 & 3.29 & 3.46 & 5.28\%\\
Openstory++ (w/o visual anno) & 3.81 & 4.01 & 3.53 & 3.62 & 4.09 & 4.12 & 11.67\% \\
Openstory++  (w/ visual anno) & 4.02 & \textbf{4.14} & \textbf{4.25} & \textbf{4.31} & \textbf{4.40} & 4.24 & \textbf{21.28\%} \\
\hline
\end{tabular}
}
\caption{Comparison results of human evaluation. The metrics are text-image alignment, style consistency, content consistency, character consistency, plot continuity, image quality, and preference, respectively. The scores for each metric have been adjusted to maintain a consistent relationship with the preference percentage.}
\label{tab:human_eval}
\end{table}

\section{Conclusion}
In summary, our work introduces significant advancements in the field of multi-modal generation. The \textbf{Openstory++ } dataset, coupled with our tailored training methodology, addresses the limitations of current image generation models by providing a rich, instance-focused resource that promotes entity consistency and narrative continuity. Furthermore, the \textbf{Cohere-Bench} benchmark framework sets a new standard for evaluation, focusing on long-context coherence and multi-turn capabilities. These contributions not only enhance the capabilities of existing models but also pave the way for future innovations in generating and interpreting complex narratives within open-domain settings.

\section{Limitations}
Although this work shows some progress, there are still some limitations. While broad, the Openstory++ dataset may not cover all possible scenarios encountered in visual storytelling. In addition, there are some errors in the algorithm of the data set, and the data set is not completely error-free, which may bring some biases to the training results.

\newpage
\small{\printbibliography{}}

\newpage

\appendix

\section{Dataset Details}
Our dataset mainly consists of 100M single-image data with instance-level annotation and 1M image sequence data with narrative content. In this section, we introduce some details of the dataset composition and the pipeline for building the dataset.
\subsection{Data Source and Data selection}
\subsubsection{Unique Data}
\begin{enumerate}
    \item \textbf{Laion-400m} This is one of the currently popular large-scale datasets, featuring open-domain large-scale images and traceable URLs. However, the original image captions vary significantly in quality. We utilized BLIP2 to re-caption these images and processed them through our single-image pipeline, resulting in a dataset format with instance-level annotations.
    \item \textbf{SA-1B} This is a large-scale image dataset that includes high-quality street view images. Due to privacy protection measures, facial data in this dataset has been blurred, which is not ideal for pretraining our instance-level model. Therefore, we only incorporated a portion of the high-quality images into our dataset.
    \item \textbf{JourneyDB} This is a large-scale dataset containing 4,429,295 high-quality Midjourney images, annotated with high-quality image captions. Since these images are generated by Midjourney, they exhibit a uniform style. We included this dataset as a separate subset and performed instance-level annotations. Due to the significant stylistic differences from real-world images, it serves as a high-quality subset for training instance-focused models.
\end{enumerate}
Once we have these image datasets, we can filter and annotate them through our single-image instance-level annotation pipeline.
\subsubsection{Continous Sequence Data}
\begin{enumerate}
    \item \textbf{Pandas-70M} This is a large-scale dataset with 70 million high-quality video-caption pairs. This dataset is generated using a multi-step process that includes video collection, captioning, and filtering. The dataset is designed to provide a comprehensive and diverse set of video-text pairs for various multimodal tasks
    \item \textbf{InternVid} This is a large-scale video-centric multimodal dataset designed to facilitate the development of powerful and transferable video-text representations for multimodal understanding and generation. This dataset contains over 7 million videos, yielding 234 million video clips accompanied by detailed descriptions totaling 4.1 billion words. The dataset is built using a scalable approach that leverages large language models (LLMs) to generate high-quality video-text pairs
\end{enumerate}
We utilized a subset of data from Pandas-70M and Intenvid, primarily sourced from naturally story-driven YouTube categories such as film and how-to, as well as subject-driven YouTube categories like pet and people. Additionally, we extracted videos corresponding to captions containing subjects from their inherent video captions. For instance, if a caption reads "A person is holding a long-haired dachshund in their arms," it indicates a video containing the subject we're interested in. From these, we curated high-quality source videos to build our dataset. Furthermore, we gathered numerous high-quality videos from narrative-focused channels with a focus on subjects, such as reality TV shows and outdoor adventures from explorer channels.

\subsection{Dataset Construction Pipeline Details}
\label{sec:pipeline}
We set up a multi-faceted filtering strategy to improve the dataset's quality. For example, aesthetic score, number of Instances, etc. 
\subsubsection{Filtering strategy}

\paragraph{Instance filtering}
We filtered out images without instances and images with more than eight subjects. The number of instances was determined by the bounding boxes annotated by YOLO-World. This is because images without instances add noise to the dataset, while images with too many instances become too complex for the model to learn effectively.

\paragraph{Aesthetic filtering}
To improve the quality of the dataset, we employed an aesthetic evaluation model to filter out images with low aesthetic scores. For unique images, we selected those with an aesthetic score greater than five, resulting in a high-aesthetic-score dataset that constitutes 15\% of the total unique image dataset. Additionally, in the image sequence dataset, we retained only origin frames with scores higher than 4.5 in our pipeline. Through human evaluation, this approach effectively filtered out transitional and blurry shots, thereby strengthening the continuity between frames
\subsubsection{Pipeline Performance and Models Details}
The performance of the entire pipeline described above is primarily constrained by the speed bottleneck of the LLM in refining the captions and the instance-level data annotation after obtaining the refined captions. We use ChatGLM3-Turbo's API as our LLM provider, and it returns our refined caption request approximately once every 5 seconds. We employed the latest YOLO-Worldv2-X and YOLO-Worldv2-L models for mixed annotation of bounding boxes in our dataset. Additionally, we utilized EfficientViT-SAM for generating instance masks within the bounding boxes. Compared to other related approaches, these models demonstrate rapid and high-performance annotation at the instance level. Their combined performance on a single A100 GPU can handle approximately 30 images per second.

\subsection{Dataset Statistics}
\begin{figure}
    \centering
    \includegraphics[width=1.0\linewidth]{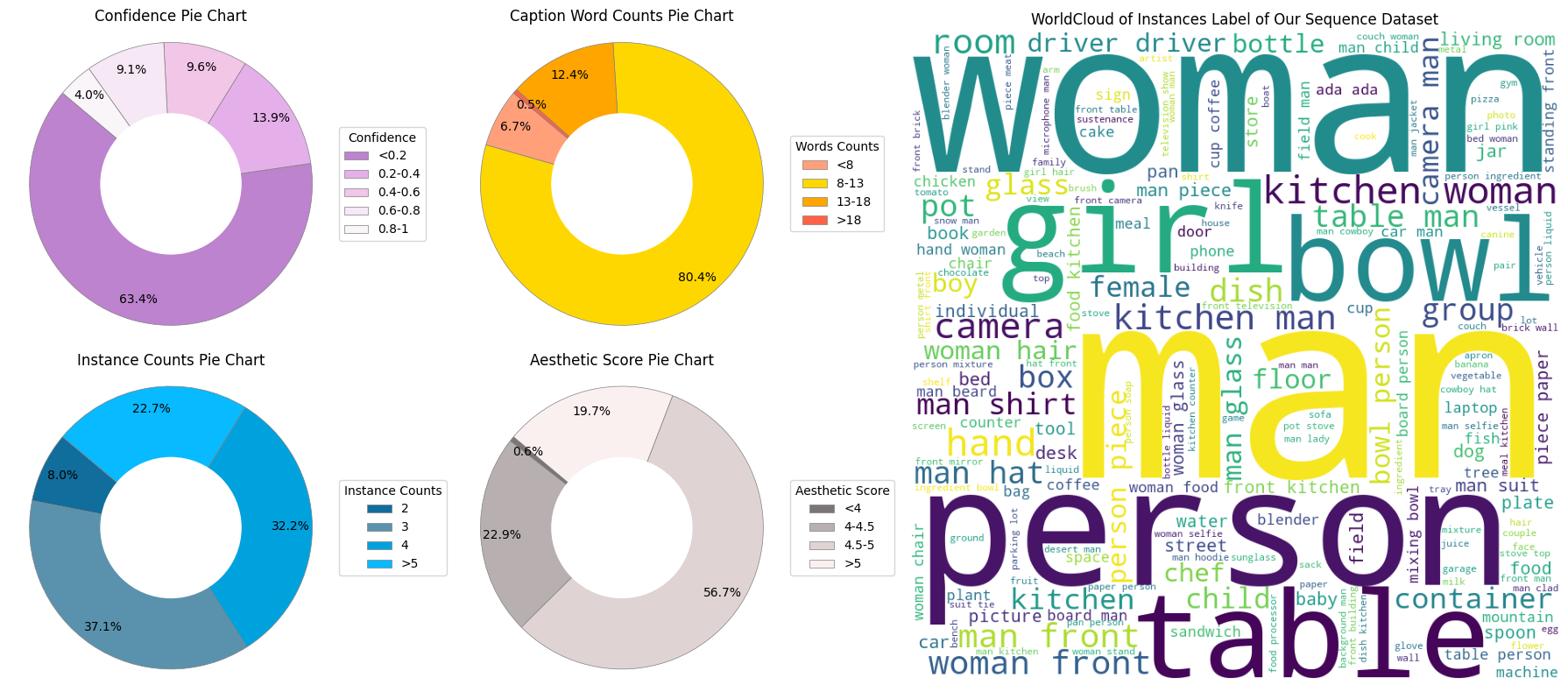}
    \caption{Some statistics about bounding box confidence, caption word count, instance count, and aesthetic score, as well as a word cloud of instance label of our sequence dataset}
    \label{fig:dataset_distribution}
\end{figure}

As illustrated in Figure~\ref{fig:dataset_distribution}, we have compiled statistics on the aesthetic scores, bounding box confidences, caption word counts, and the number of entities per image in our sequence data. Additionally, we present a word cloud to display the information on labels within our bounding boxes, which intuitively shows the diversity of our data.

Besides, It is worth noting that when we mentioned filtering out images with aesthetic scores lower than 4.5, this filtering was done at the source image stage. Since our final target images were resized to match the typical square size used by encoders, the aesthetic scores exhibited some variations upon recalculation.

\section{Experiments}
\label{sec:exp}

In this section, we start by describing our experimental settings and then compare them with other models. Additionally, we present results for ablation experiments to prove the effectiveness of our dataset construction pipeline modules. 

\begin{figure}
    \centering
    \includegraphics[width=0.95\linewidth]{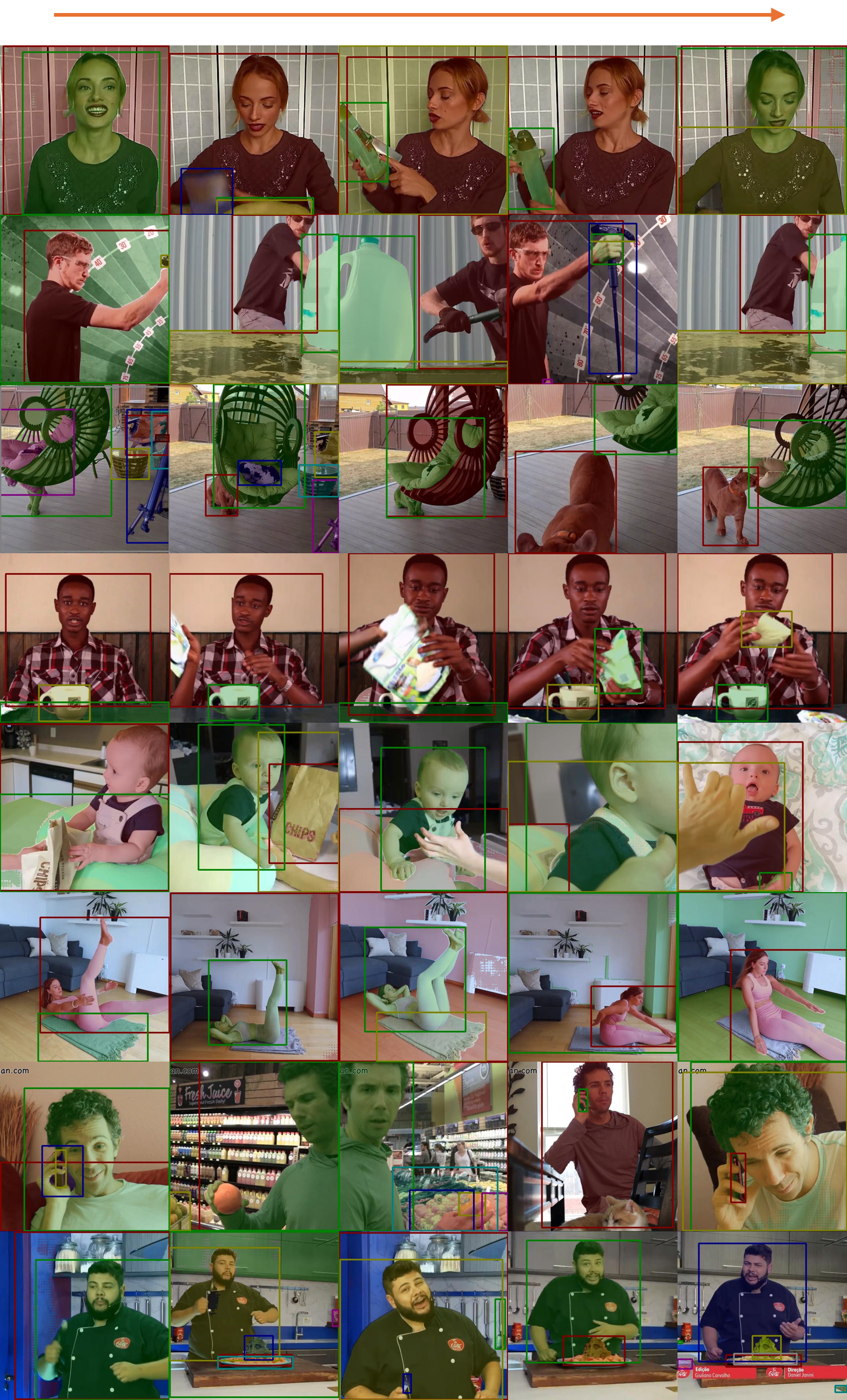}
    \caption{Some single instance sample from our dataset}
    \label{fig:single_instance}
\end{figure}

\begin{figure}
    \centering
    \includegraphics[width=0.95\linewidth]{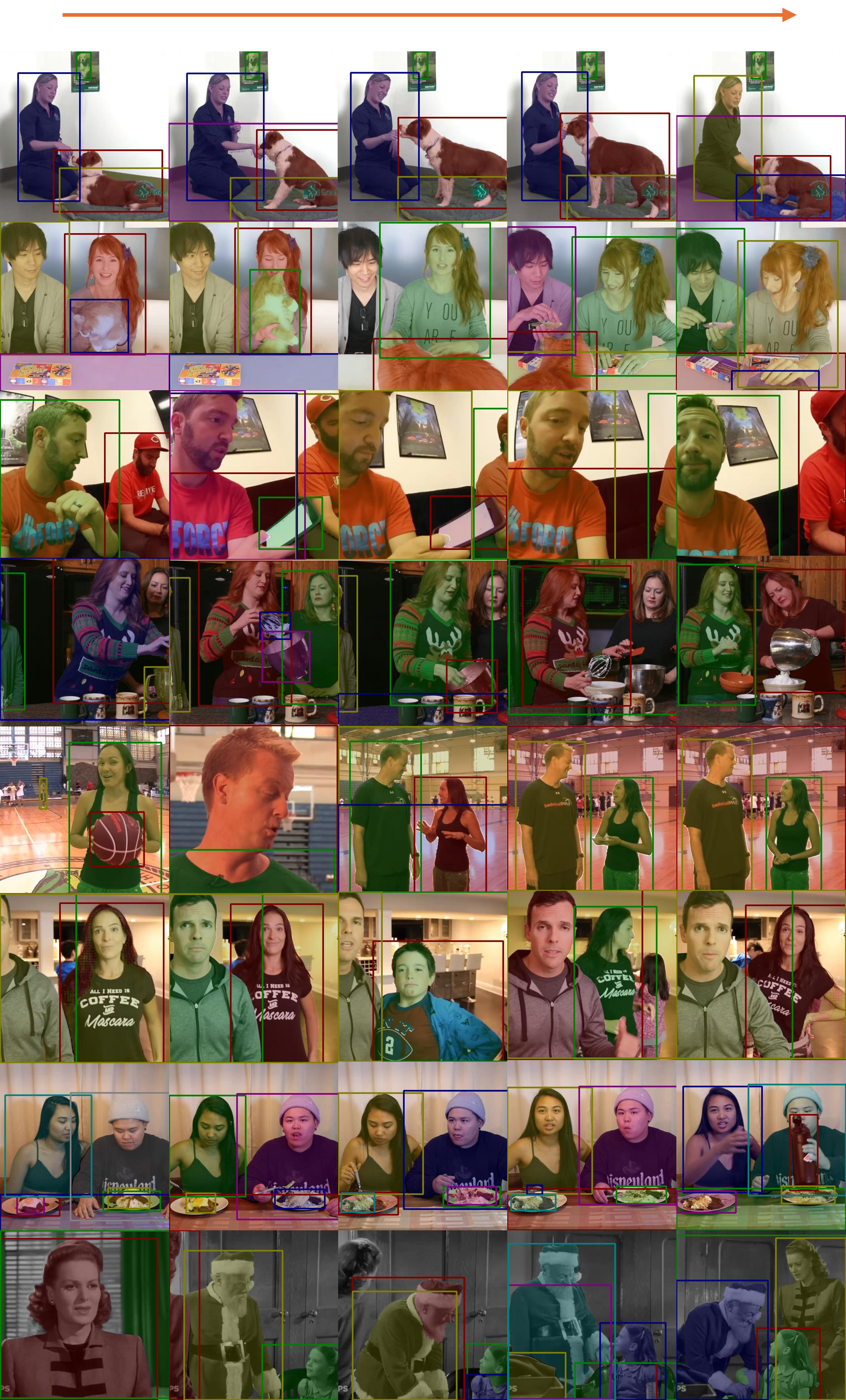}
    \caption{Some multi instance sample from our dataset}
    \label{fig:mulit_instance}
\end{figure}

\subsection{Experimental Settings}

\paragraph{Validation Dataset Set}
We selected two subsets from the OpenStory++ dataset for our evaluation: the single dataset~\ref{fig:single_instance} and the multi dataset~\ref{fig:mulit_instance}. These subsets represent scenes with single subjects and multiple subjects, respectively. Our selection process involved manual evaluation. Specifically, we categorized the collected dataset by video ID into different stories. We then observed each story to ensure that all frames within a story contained the same subject performing different actions. Only those stories that met this criterion were included in our evaluation data. The single dataset typically consists of stories with a main instance in each frame, while the multi dataset includes stories with multiple main instances. This careful curation ensures that the evaluation data accurately represents the complexity and variety of real-world scenarios, allowing us to robustly assess the generative capabilities of the models.

\paragraph{Cohere-Bench Task Design} Once we have a validation set, we designed two tasks within the Cohere-Bench framework to evaluate the generative capabilities of current large multi-modal models: story generation and story continuation. For the story generation task, we provided the multi-modal models with a text prompt to generate an image for the first scene. Subsequently, the models used the generated image and a new text prompt for the second scene to generate the corresponding image, continuing this pattern for subsequent scenes. In the story continuation task, we initially provided both the image of the first scene and the text prompt for the second scene, aiming to generate the image for the next scene. For both tasks, each story comprised 2-5 scenes.

\begin{figure}
    \centering
    \includegraphics[width=1\linewidth]{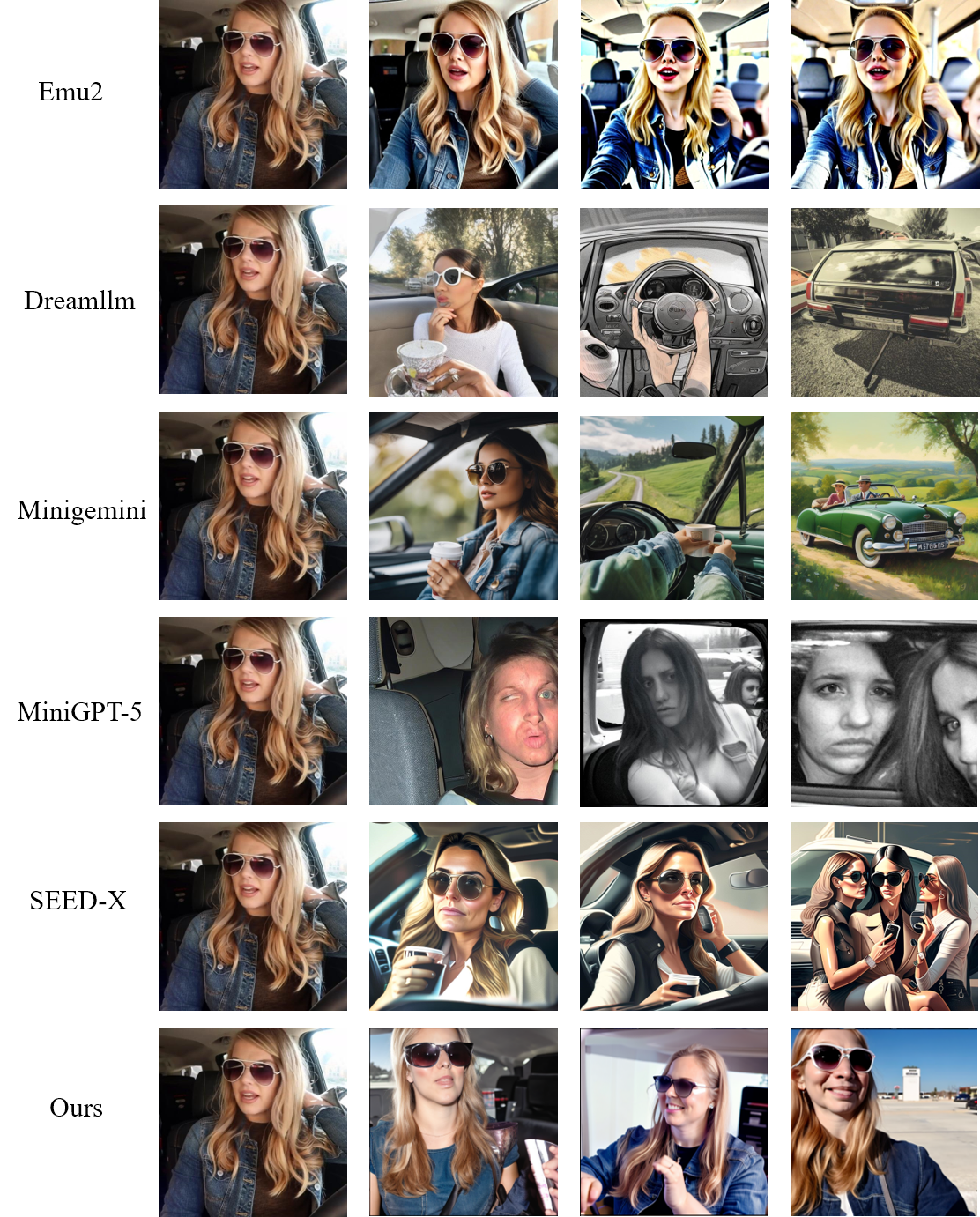}
    \caption{Here are the results of evaluating different models. The first image is the reference image. We used the captions of the last three images from our validation set as part of the prompt input to each model. These three captions are: "a female in sunglasses is driving a vehicle while sipping from a cup," "a female operator of a vehicle with her hand0 and hand1 placed firmly on the steering wheel," and "woman0 is capturing a self-portrait in front of a structure."}
    \label{fig:good_case}
\end{figure}

\paragraph{Test Details} 
We tested these tasks on many models: GPT4-V, Seed-X, Emu2, DreamLLm, MiniGemini, MiniGPT-5 and our model. For each model, we used the official open-source inference settings as default parameters. For models that did not support simultaneous text and image input to generate images, we manually adjusted the input and generation methods without making any other modifications. All inferences were performed on a single A100 GPU.For each model, the text prompt followed the format "Generate an image + {caption for the next scene}", with MiniGPT-5 using its default prompt description as input. The image prompt for each scene was the generated image of the previous scene. In addition, when we evaluate other models, we will delete the dino alignment index in caption, such as 0 in "woman0". One results of a model comparison can be referred to Figure ~\ref{fig:good_case}. We assessed the generative quality of each model using the metrics introduced in the Section ~\ref{sec:cohere-bench}.

\paragraph{Cohera-Benchmark Evaluation}
After we have the experimental results obtained in section~\ref{sec:exp}, based on which, we can introduce in detail how we evaluate each model, which is mainly divided into six parts: Semantic Alignment, Background Consistency, Style Consistency, Instance Consistency, Instance Integerity, and BLEU4.

\begin{itemize}
    \item \textbf{Semantic Alignment} Since we use the dataset's  captions as part of the prompts, we leverage CLIP-ViT-B-32 to encode the given captions and the model-generated images from the validation set. We then compute the similarity between their feature vectors. This similarity serves as our Semantic Alignment score.
    \item \textbf{Background Consistency} We employ an open-domain detector, Yolo-World, to detect all instances in the images, focusing on generic classes like characters and animals. After masking these instances and inpainting the masked regions, we obtain background images. By encoding the generated image sequences with CLIP-ViT-B-32, we calculate the similarity between the background of the reference image (the first image of each scene in the validation set) and the backgrounds of subsequent images. The average similarity score gives us the Background Consistency score.
    \item \textbf{Style Consistency} We use dino-vits16 to encode all images in the generated sequence. We then compute the feature similarity between consecutive frames and take their average to obtain the Style Consistency score.
    
    \item \textbf{Instance Integrity} Similar to the Background Consistency calculation, we use Yolo-World to obtain bounding boxes for all instances. Using the bounding box of characters from the reference image, we encode these regions with dino-vits16 to get the base features. For the subsequently generated images, we encode the bounding boxes of characters to obtain their visual features. We then construct a similarity matrix between these features and the base features. Specifically, 
    \begin{equation}
    \mathrm{Instance~Integrity~Score} = \frac{\sum_k \mathrm{Similarity}(f_{i_k}^\mathrm{current}, f_{j_k}^\mathrm{base})}{\mathrm{len}(\mathbf{F}_\mathrm{base})}
    \end{equation}
    where $(i_k, j_k)$ are the optimal matching index pairs found using the Hungarian algorithm on the cost matrix $C$.
    \item \textbf{BLEU4} The BLEU4 score is calculated by comparing the caption provided in the prompt with the caption generated by BLIP2 for the model-generated image.
\end{itemize}
The similarity between all feature vectors is calculated using the following formula:
\begin{equation}
\text{Similarity} = \frac{1}{N} \sum_{i=1}^{N} \mathbf{f}_i^\text{feature1} \cdot (\mathbf{f}_i^\text{feature2})^T
\end{equation}
where \( \mathbf{f}_i^\text{feature1} \) and \( \mathbf{f}_i^\text{feature2} \) are the feature vectors from feature set 1 and feature set 2, respectively, and \( N \) is the total number of feature vectors. The dot product \( \mathbf{f}_i^\text{feature1} \cdot (\mathbf{f}_i^\text{feature2})^T \) calculates the similarity between each corresponding pair of feature vectors, and the mean of these similarities provides the overall similarity score.

\paragraph{Human Evaluation}
To better assess human preferences for the generated stories, we conducted a human evaluation using several criteria: Align, Style, Consistency, Character, Plot Continuity, Image Quality, and Preference Evaluation.

\begin{itemize}
    \item \textbf{Align} Participants were given each scene's image and the corresponding text prompt. They rated the alignment between the image and the text on a scale of 1 to 5, with 1 being the lowest and 5 the highest. The participants were not informed about which model generated the images.
    \item \textbf{Style} Participants were shown the image sequences for each story and rated the stylistic similarity between the images in the sequence on a scale of 1 to 5. The rating scale was the same, and the participants were not aware of the model origins.
    \item \textbf{Consistency} Different story text and image sequences were presented to participants, who rated the consistency between the story text and the generated images on a scale of 1 to 5. The identity of the generating model was kept anonymous.
    \item \textbf{Character}
    For each story, participants evaluated the consistency of the same subject across different images. The primary focus was on whether the subject remained consistent with previous appearances. Ratings were given on a scale of 1 to 5, without knowledge of the image's source model.
    \item \textbf{Plot Continuity} To ensure that the generated scenes were logically connected, participants rated the logical coherence between different scenes within each story based on both text and images. The rating scale was 1 to 5, and participants were blinded to the model generating the images.
    \item \textbf{Image Quality} The quality of the images is crucial for story generation. Participants evaluated the aesthetic quality of the generated images on a scale of 1 to 5, again without knowing the source model.
    \item \textbf{Preference Evaluation} Participants were asked to provide an overall assessment of the images generated for each story. They viewed image sequences of the same story generated by different models, without knowing which model produced each sequence, and selected their preferred image sequence based on personal preference.
\end{itemize}

\subsection{Ablation Studies of Pipeline}

\begin{table}[ht]
    \centering
    \begin{tabular}{c c c}
         \toprule
         \textbf{} & \makecell{Semantic\\Alignment}~$\uparrow$ & Perplexity~$\downarrow$ \\
         \midrule
         BLIP2 Captions & 0.228 & 38.024 \\
         Refined Captions & \textbf{0.262} & \textbf{29.045} \\
         \bottomrule
    \end{tabular}
    \caption{Performance comparison of BLIP2 and Refined Captions in terms of Semantic Alignment and Perplexity.}
    \label{tab:caption_ablation}
\end{table}

To validate the effectiveness of LLM-refined captions, we measured the semantic similarity between the captions and images both before and after refinement. Additionally, we assessed the perplexity of the captions prior to refinement. The results are shown in Table ~\ref{tab:caption_ablation}. It can be seen that using LLM to refine the caption can make the caption more narrative and more accurate in describing the image.

\end{document}